\pdfoutput=1

\documentclass[11pt]{article}

\usepackage[]{EACL2023}

\usepackage{inconsolata}

\usepackage{times}
\usepackage{latexsym}

\usepackage[T1]{fontenc}

\usepackage[utf8]{inputenc}

\usepackage{microtype}

\usepackage{microtype}
\usepackage{graphicx}
\usepackage{subfigure}
\usepackage{booktabs} 

\usepackage{amsmath,amsfonts,bm}

\def\eqref#1{equation~\ref{#1}}

\def\1{\bm{1}}

\DeclareMathAlphabet{\mathsfit}{\encodingdefault}{\sfdefault}{m}{sl}
\SetMathAlphabet{\mathsfit}{bold}{\encodingdefault}{\sfdefault}{bx}{n}

\usepackage{CJKutf8}

\usepackage{hyperref}
\usepackage{url}
\usepackage{graphicx}
\usepackage{booktabs}
\usepackage{multirow, multicol}
\usepackage{diagbox}

\usepackage{tabularx}
\usepackage{xspace}
\usepackage{amsmath}
\usepackage{float}
\usepackage{xcolor}
\usepackage{tablefootnote}

\usepackage{pifont}
\usepackage{amssymb}

\usepackage{enumitem}

\newcommand\chin[1]{\begin{CJK*}{UTF8}{gbsn}#1\end{CJK*}}

\newcommand{\TODO}[1]{{\color{blue} TODO: #1}}

\title{Contextual Semantic Parsing for Multilingual Task-Oriented Dialogues}

\author{Mehrad Moradshahi$^{1}$ \quad Victoria Tsai$^{2}$ \quad  Giovanni Campagna$^{1}$ \quad  Monica S. Lam$^{1}$ \\
  $^1$ Stanford University, Stanford, CA \\
  $^2$ Meta, Menlo Park, CA \\
  \texttt{\{mehrad,gcampagn,lam\}@cs.stanford.edu} \\
  \texttt{\{vitsai\}@meta.com} \\
}

\begin{document}
\maketitle

\begin{abstract}

Robust state tracking for task-oriented dialogue systems currently remains restricted to a few popular languages.
This paper shows that given a large-scale dialogue data set in one language, we can automatically produce an effective semantic parser for other languages using machine translation.  We propose automatic translation of dialogue datasets with alignment to ensure faithful translation of slot values and eliminate costly human supervision used in previous benchmarks. We also propose a new contextual semantic parsing model, which encodes the formal slots and values, and only the last agent and user utterances. We show that the succinct representation reduces the compounding effect of translation errors, without harming the accuracy in practice.

We evaluate our approach on several dialogue state tracking benchmarks. On RiSAWOZ, CrossWOZ, CrossWOZ-EN, and MultiWOZ-ZH datasets we improve the state of the art by 11\%, 17\%, 20\%, and 0.3\% in joint goal accuracy. We present a comprehensive error analysis for all three datasets showing erroneous annotations can lead to misguided judgments on the quality of the model. 

Finally, we present RiSAWOZ English and German datasets, created using our translation methodology. On these datasets, accuracy is within 11\% of the original showing that high-accuracy multilingual dialogue datasets are possible without relying on expensive human annotations. We release our datasets and software open source.\footnote{\url{https://github.com/stanford-oval/dialogues}}

\end{abstract}
\section{Introduction}
\label{intro}

Tremendous effort has gone into the research and development of task-oriented dialogue agents for English and a few other major languages in recent years. A methodology that can transfer the effort to other languages automatically will greatly benefit the large population of speakers of the many other languages in the world. 


Underlying an effective TOD agent is \textit{dialogue state tracking}, the task of predicting a formal representation of the conversation sufficient for the dialogue agent to reply, in the form of slots and values. However, DST currently remains restricted to a few popular languages~\cite{razumovskaia2021crossing}.  Traditional DST agents require large hand-annotated \textit{Wizard-of-Oz}~\cite{kelley1984iterative} datasets for training, which are
prohibitively labor-intensive to produce in most languages~\cite{gunasekara2020overview}. Large, multi-domain WOZ datasets are only available in English and Chinese~\cite{quan2020risawoz,ye2021multiwoz}.

\noindent The contributions of this paper are as follows: 

    {\bf 1. We propose an automatic technique to build multilingual data sets using machine translation.} Machine translation has been shown effective for localizing question-answering agents~\cite{moradshahi-etal-2020-localizing}. It shows that for open ontology datasets, we need to use an alignment model to properly translate entities in the source language to entities in the target language. This paper shows that alignment is necessary even for closed ontology datasets and dialogues. 
    
    Furthermore, we improve alignment to address these challenging issues we discovered unique to dialogues: (1) Translation errors accumulate and can prevent a correct parse for the rest of the dialogue; (2) There are logical dependencies between slot values across different turns; (3) Utterances are generally longer and more complex carrying multiple entities. We found that alignment improves the accuracy on the RiSAWOZ benchmark by 45.6\%. This technique eliminates the cost of human post-editing used on all previous translation benchmarks, and can improve machine translation quality on other tasks too. 

    Using this methodology, we automatically translate the RiSAWOZ dataset to English and German, creating RiSAWOZ-EN-auto and RiSAWOZ-DE-auto datasets respectively. 
    
    
    
    {\bf 2. We show that the accumulation of translation and annotation errors across turns can be mitigated with a \textit{Contextual Semantic Parsing} (CSP) model for state tracking.} 
    We propose a BART-CSP model, a seq-to-seq based on BART, that encodes the belief state, and the last agent and user utterances, rather than the full history of utterances.

    BART-CSP improves SOTA on RiSAWOZ~\cite{quan2020risawoz} and CrossWOZ~\cite{zhu2020crosswoz}, two large-scale multi-domain WoZ dialogue datasets, by 10.7\% and 17\% in Joint Goal Accuracy(JGA). Notably, BART-CSP is more effective on translated data as evident by bigger performance improvement: on RiSAWOZ-EN-auto and RiSAWOZ-DE-auto datasets, automatically translated versions of RiSAWOZ, BART-CSP improves SOTA by 32.4\% and 52.5\%.
\section{Related Work}
\label{related_work}

\subsection{Dialogue State Tracking}
Dialogue state tracking (DST) refers to the task of predicting a formal state of a dialogue at its current turn, as a set of slot-value pairs at every turn. State-of-the-art approaches apply large transformer networks~\cite{peng2020soloist, hosseini2020simple} to encode the full dialogue history in order to predict slot values. Other approaches include question-answering models~\cite{gao2019dialog}, ontology matching in the finite case~\cite{lee-etal-2019-sumbt}, or pointer-generator networks~\cite{Wu2019May}. Both zero-shot cross-lingual DST transfer~\cite{ponti2018adversarial, chen2018xl} and multilingual knowledge distillation~\cite{hinton2015distilling,tan2019multilingual} have been investigated; however, training with translated data is the dominant approach, outperforming zero-shot and few-shot methods.

\subsection{Contextual Semantic Parsing}
Alternatively to encoding the full dialogue history, previous work has proposed including the state as context~\cite{lei2018sequicity, heck2020trippy, ye2021slot} together with the last agent and user utterance. Recently, \newcite{cheng-etal-2020-conversational} proposed replacing the agent utterance with a formal representation as well. Existing models rely on custom encoder architectures and loss functions for the state~\cite{heck2020trippy}. Our formulation of CSP is different since we encode the formal dialogue state directly as text, which simplifies the architecture and makes better use of the pretrained model's understanding of natural text.

Previous work also applied rule-based state trackers that compute the state based on the agent and user dialogue acts~\cite{schulz2017frame, zhong2018global, zhu2020crosswoz}. Such techniques cannot handle state changes outside of a state machine defined ahead of time and do not achieve state-of-the-art accuracy on WOZ dialogues.

\subsection{Multilingual Dialogues}
Several multilingual dialogue benchmarks have been created over the past few years. Dialogue State Tracking Challenge (DSTC) has released several datasets~\cite{kim2016fifth,hori2019overview,gunasekara2020overview}, covering only a few domains and languages. CrossWOZ~\cite{zhu2020crosswoz} and RiSAWOZ~\cite{quan2020risawoz} are Chinese datasets collected through crowdsourcing. BiToD~\cite{lin2021bitod} uses a dialogue simulator to generate dialogues in English and Chinese, then uses crowdsourcing to paraphrase entire dialogues. All these approaches use crowdworkers in one or multiple stages of data collection which is costly and human errors degrade quality. Automatic creation of affordable high-quality dialogue datasets for other languages still remains a challenge~\cite{razumovskaia2021crossing}. 

\section{Task Setting}
We are interested in the \textit{dialogue state tracking} task, in which the goal is to predict a formal representation of a conversation up to a certain point, also known as \textit{belief state}, consisting of the slots that were mentioned in the dialogue and their value. At the beginning of the conversation, the belief state is empty, and it grows as the conversation progresses, accumulating the slots that were mentioned across all turns prior.

Formally, the problem is formulated given a predefined set of \textit{slots} $s_1, s_2, \ldots s_n$ (such as ``restaurant name'', ``restaurant food'', etc.). Each slot has one value taken from the \textit{ontology} $v_1, v_2, \ldots, v_n$. The ontology contains the legitimate values for the slot from the database (i.e. the list of restaurant names or restaurant cuisines), as well as the special values ``none'' indicating the slot was not mentioned, and ``dontcare'' indicating the slot was explicitly mentioned by the user but the user has no preference.

Therefore, given a partial conversation history composed of turns $x_1, x_2, \ldots, x_t$, where each turn consists of an agent response ($a_t$) and user utterance ($u_t$), the task is to predict the belief state:
\resizebox{1.0\hsize}{!}{$b_n(x_1, \ldots, x_t) = \{ s_1 = v_{1,t}, s_1 = v_{2,t}, \ldots, s_n = v_{n,t} \}$}
where $n$ is the number of slots and $v_{i,t}$ is the value of the slot $s_i$ up to turn $t$ of the conversation. Note that the slot could be mentioned at turn $t$, or in any of the turns before.

\begin{figure}
\centering
\includegraphics[width=\linewidth]{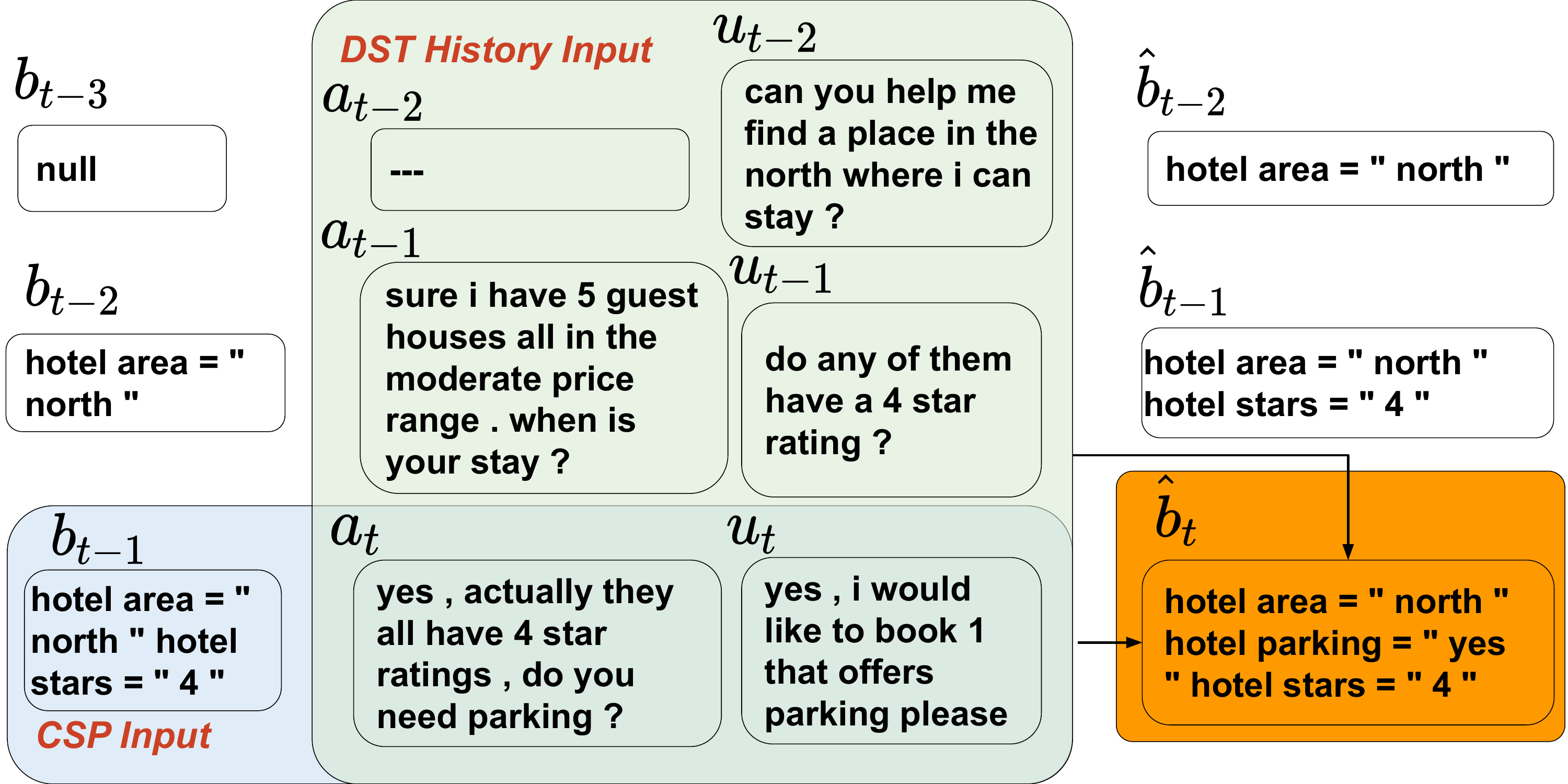}
\caption{Visualization of the differences between inputs of CSP and traditional DST models. While the latter encodes the full history of the dialogue, CSP only encodes the current state and turn. $b$, $a$, and $u$ indicate the dialogue state, agent utterance, and user utterance, respectively. $t$ is the turn number. \^~indicates a predicted dialogue state.}
\label{fig:cspdst}
\vspace{-1em}
\end{figure}

\section{Reducing Translation Noise}
Key to the success of the dialogue state tracking is a precise and consistent annotation of the belief state at every turn. This is challenging in the multilingual setting, where we apply automatic translation to produce new training sets for DST: (1) the translation can be ungrammatical or incorrect, and it can introduce spurious information, and (2) the new annotation (in the target, translated ontology) can diverge from the translation of the sentence, such as referring to the same value in different ways in the target language.

We refer to the error introduced in translation collectively as \textit{translation noise}. We posit that translation noise is the reason for poorer performance in existing translated dialogue datasets, compared to the same model on the dialogue dataset in the source language. In this section, we describe our methods to reduce translation noise.

\subsection{Alignment}
\label{sec:align}
A major source of translation noise is due to mismatches between the translation of an entity by itself and in sentence.
For instance, given a Chinese sentence that refers to an utterance containing the word ``aquarium'' (\chin{水族馆}), incorrect translation may result in ``ocean museum'' in the English utterance, which does not match the slot value ``aquarium'' in the ontology and annotation anymore. Slot values may also get dropped or transliterated.

\begin{figure}
\centering
\includegraphics[width=\linewidth]{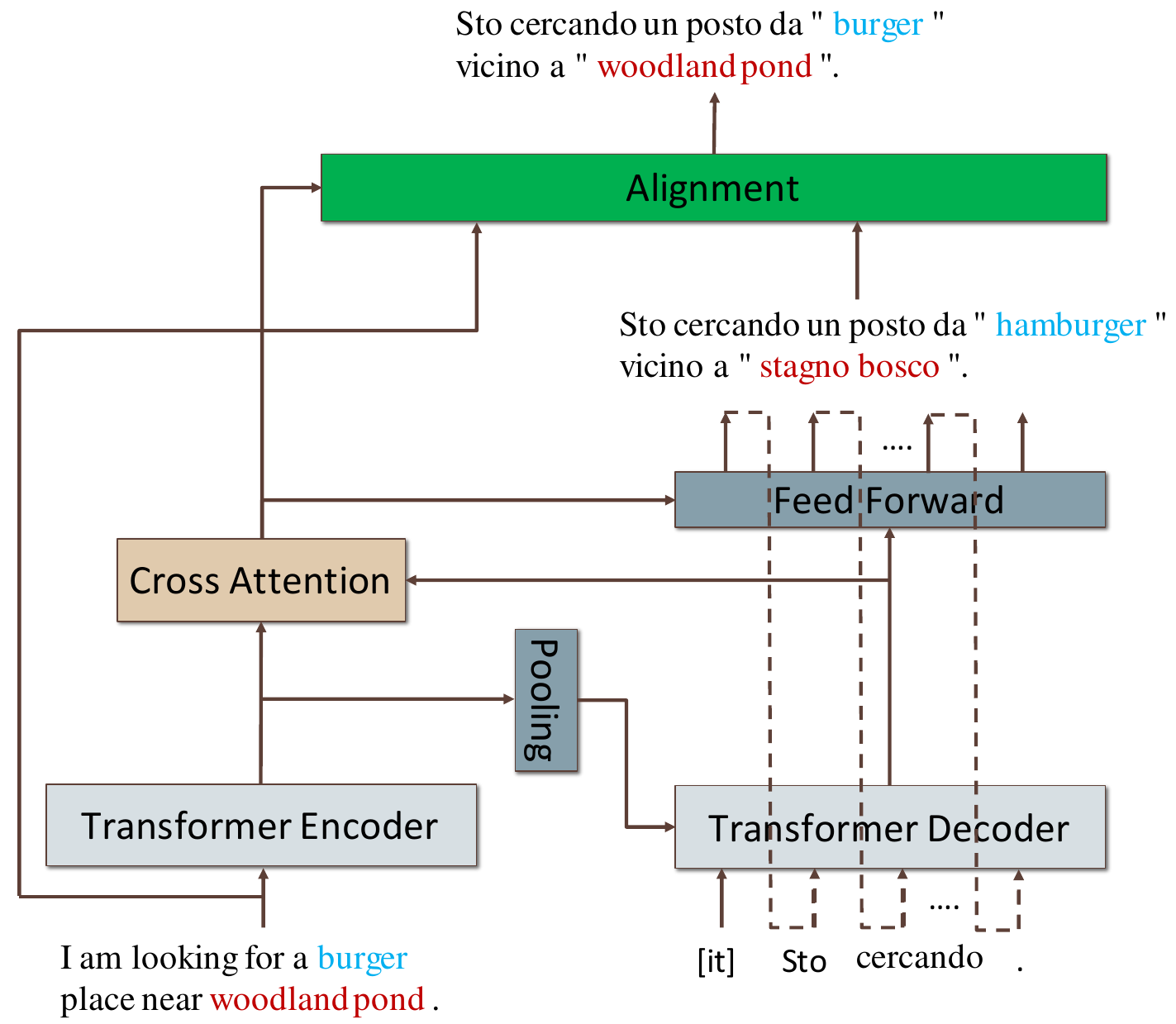}
\caption{Example of the translation and alignment process. The encoder-decoder cross-attention weights are used to achieve word alignment.}
\label{fig:translation}
\end{figure}

\textit{Translation with alignment} was previously proposed by~\newcite{moradshahi-etal-2020-localizing} to localize open-ontology multilingual semantic parsing datasets. Token alignments, obtained from cross-attention weights of the neural model, are used to track position of entities during translation so they can be replaced afterwards with local entity values. Figure~\ref{fig:translation} shows the translation and alignment process for an example input.

We show that alignment is useful also for a finite (closed) ontology in a dialogue setting. The dialogue setting is more challenging since the replacement with local entity values must be consistent across turns and dependent slots - slots that their values are logically dependent on each other. For instance, the corresponding price range for a \textit{fast food} restaurant should be \textit{cheap}, or a speaker looking for an attraction to go to with his girlfriend wants a place where \textit{best-for-crowd} = ``lover's date''. Furthermore, utterances are generally longer and more complex containing multiple entities.

We have made several changes in alignment to address these issues:
\begin{enumerate}
    \item 
We use a dictionary constructed from the dataset's ontology for translating the dependent slots to ensure relations are preserved. For all other slots, we randomly replace them with values from the target language ontology similar to previous work.

\item
In previous work, quotation marks were used to mark the boundary of entities and to retrieve alignment between tokens in the input and output. We found the translation of quotation marks to be inconsistent. Instead, we omit those marks before translation and purely rely on cross-attention between subwords to compute alignment. 

\item
We observed alignment does poorly on digits and often misplaces them in the output. We use string matching to retrieve spans for numbers, dates, and time slots if present in the output and omit alignment if successful.
\item
Dialogues contain longer utterances with multiple entities per turn. We found breaking down utterances into individual sentences before translation, significantly improves the quality of outputs when there are fewer entities to align. 
\end{enumerate}

To measure the effect of our changes, we translate MultiWOZ English dev set to Chinese with the new and previous alignment methods using MBART-MMT (see Section~\ref{sec:impl} for details). The new approach improves BLEU score by 8 points and JGA by 5\% on MultiWOZ-ZH dev set.

\subsection{Contextual Semantic Parsing}

Previous work on dialogue state tracking encodes all or several turns up to the current one using a neural model, which then predicts the value for each slot. Hence, the input of the model consists exclusively of natural language, and grows as the conversation grows, accumulating any translation noise.

At the same time, we observe that the belief state at turn $t$, $b_t$, can be computed from the belief state at turn $t-1$ and the slots mentioned in the utterances at turn $t$. Hence, we propose to use a \textit{contextual semantic parser} (CSP) for dialogue state tracking that computes $P(b_t | a_t; u_t; b_{t-1})$. The CSP model is applied to the dialogue-state tracking task by iteratively predicting the belief state of all turns, starting from $b_0$, the initial state consisting of all empty slots.

The CSP formulation condenses the dialogue history into a formal, fixed-length representation. Because the representation does not grow with the dialogue, it does not suffer from accumulation of translation noise.


Our CSP model is based on Seq2Seq Transformer models BART~\cite{lewis2019bart} for English and MBART~\cite{liu2020multilingual} for all others. Here we refer to them as CSP-BART for simplicity. 

\begin{figure}
\centering
\includegraphics[width=\linewidth]{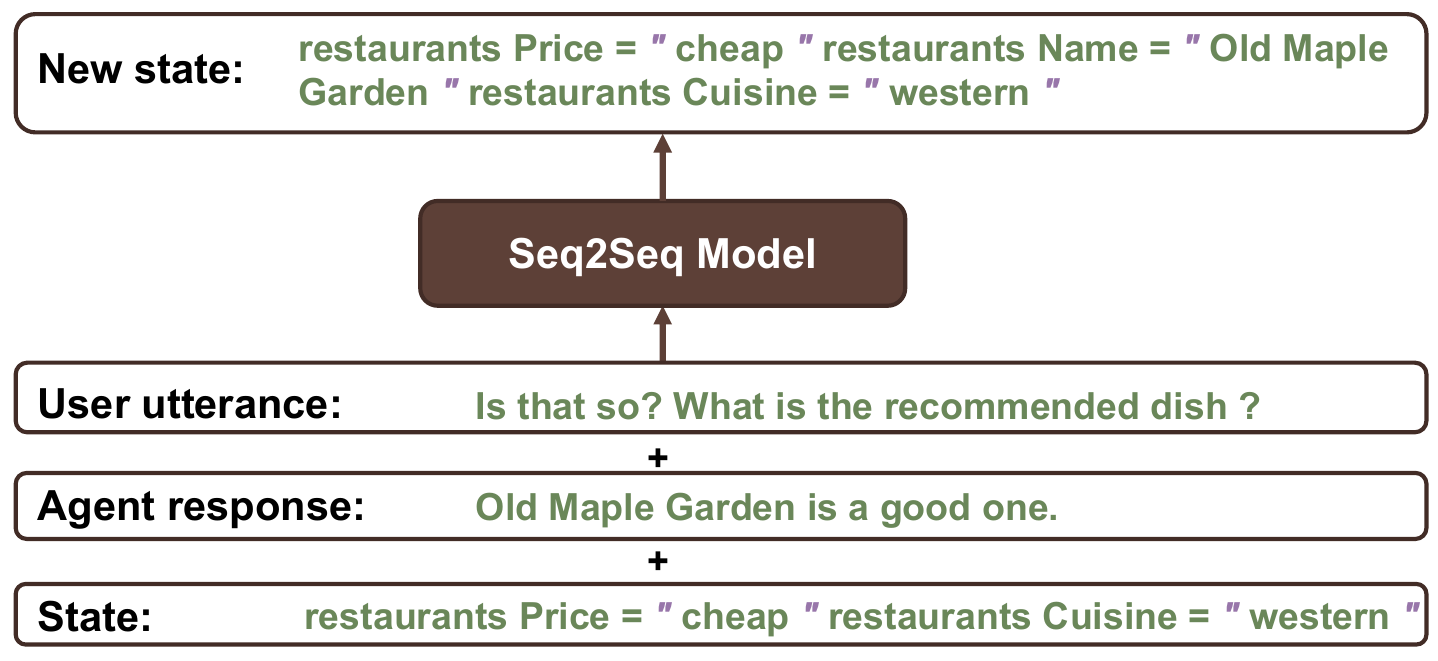}
\caption{Contextual Semantic Parser model. We use BART for English and MBART for Chinese and German as the seq2seq model.}
\label{fig:dialogue-loop}
\end{figure}

The model encodes the belief state as a textual sequence of slot names and slot values. This encoding is concatenated to the agent utterance and user utterance, and fed to the model to predict the belief state at the end of the turn (Fig.~\ref{fig:dialogue-loop}). Similar to~\cite{yang2021ubar}, we encode the belief state directly as text, which simplifies the architecture and leverages the pretraining of BART.

\section{Experiments}
\label{sec:experiments}




Our experiments are designed to answer the following questions: 1)  How well does CSP perform on WOZ datasets compared to DST models that encode the full conversation history? 2) Is our approach effective to reduce the translation noise?

\subsection{Datasets}
\label{sec:datasets}

We evaluate our models on the RiSAWOZ~\cite{quan2020risawoz}, MultiWOZ~\cite{budzianowski2018large, eric2019multiwoz}, and CrossWOZ~\cite{zhu2020crosswoz} datasets and their available translated versions: MultiWOZ Chinese~\cite{li2021multi-domain}, CrossWOZ English~\cite{li2021multi-domain}. These particular datasets were chosen because they are large Wizard-of-OZ dialogue datasets and therefore more natural and representative of task-oriented dialogues. Additionally, we use our methodology to create the RiSAWOZ English and German datasets.

RiSAWOZ~\cite{quan2020risawoz} is a Chinese WOZ dataset of 11k annotated dialogues and over 150k utterances spanning the 12 domains of attraction, restaurant, hotel, flight, train, weather, movie, TV, computer, car, hospital, and courses. Dialogues are formed from both single-domain and multi-domain goals, and annotated with dialogue states, dialogue acts, and coreference clusters. 

MultiWOZ is an English-language WOZ dataset of 10k single- and multi-domain dialogues spanning the following 7 domains: taxi, restaurant, hotel, attraction, train, police, hospital. Following prior work with this dataset~\cite{lee-etal-2019-sumbt,kim2019efficient}, we drop hospital and police from the training set as they are not included in the validation and test set. After the release of MultiWOZ 2.0~\cite{budzianowski2018large}, later iterations~\cite{eric2019multiwoz, zang2020multiwoz, han2020multiwoz} corrected some of the misannotations.

CrossWOZ is a Chinese-language WOZ dataset of 6k dialogues and over 102k utterances spanning the same 5 domains as the MultiWOZ validation and test sets: hotel, restaurant, attraction, metro, and taxi. CrossWOZ dialogues are annotated with dialogue states and dialogue acts, and average over 3.24 domains per dialogue, as opposed to the 1.80 of MultiWOZ. 

For DSTC-9, Google NMT was used to translate MultiWOZ 2.1 to Chinese  and CrossWOZ to English~\cite{gunasekara2020overview,li2021multi-domain}. To ensure translations of slot values in the dialog are faithful to the ontology, they replace them with their translations from a dictionary before feeding it to the NMT. This approach creates mixed-language sentences, shifting input distribution away from what public NMTs have been trained on, thus reducing quality~\cite{moradshahi-etal-2020-localizing}. Note that human translators were employed to proofread the translations and check certain slots to ensure values are correctly translated. 

Table~\ref{tab:train_stats} shows a comparison of statistics for the training split of datasets used or created in this work. All the datasets have a closed ontology: the same entities appear in the train, validation, and test sets.


\subsection{Models}
We compare BART-CSP results against SOTA for each dataset. Models include the following:

\begin{itemize}[leftmargin=*,noitemsep]
\item \textbf{TRADE}~\cite{Wu2019May} uses a sequence-to-sequence architecture that encodes all utterances in the dialogue. It uses a pointer-generator to output the value of each slot. 
\item \textbf{MLCSG}~\cite{quan-xiong-2020-modeling} extends TRADE by improving modeling of long contexts through a multi-task learning framework.
\item \textbf{SOM}~\cite{kim2019efficient} considers dialogue state as an explicit fixed-sized memory and uses state operation to selectively update slot values at each turn.
\item \textbf{SUMBT}~\cite{lee-etal-2019-sumbt} applies a BERT encoder to each utterance on the dialogue and a recurrent network to compute a representation of the whole dialogue, which is then matched against the ontology of each slot. 
\item \textbf{MinTL}~\cite{lin2020mintl} uses more recent pretrained models such as T5~\cite{raffel2019exploring} and BART as the dialogue utterance encoder and builds an end-to-end dialogue model jointly learning dialogue state tracking, policy, and natural language generation tasks.
\item \textbf{STAR}~\cite{ye2021slot} STAR uses two BERT models for encoding context and slot values. Additionally, they use a slot self-attention mechanism that can learn the slot correlations automatically. They use as input both the previous belief state and history of dialogue. 

\end{itemize}



\subsection{Implementation Details}
\label{sec:impl}

BART-CSP is implemented using the Huggingface~\cite{wolf2019huggingface} and GenieNLP~\cite{geniepldi19} libraries. We use the available open-source code for the other models. Hyperparameters for BART-CSP are discussed below; hyperparameters for other models are taken from the respective papers.

For semantic parsing we used \textit{bart-base} (\textasciitilde 139M parameters) for English and \textit{mbart-large-50} (\textasciitilde611M parameters) model for other languages. For translation we used \textit{mbart-large-50-many-to-many-mmt} (\textasciitilde611M parameters) which can translate directly between any pairs of 50 languages it supports. All models use a standard Seq2Seq architecture with a bidirectial encoder and left-to-right autoregressive decoder. All the models are pretrained using text denoising objective. \textit{mbart-large-50-many-to-many-mmt} is additionally finetuned to do translation. BART uses BPE~\cite{gage1994new} to tokenize the input sentences whereas MBART uses sentence-piece model~\cite{kudo2018sentencepiece}. We used Adam~\cite{kingma2014adam} as our optimizer with a learning rate of $1 \times 10^{-5}$ and used transformer warmup schedule~\cite{popel2018training} with a warmup of 80. These hyperparameters were chosen based on a very limited hyperparameter search on the validation set. For the numbers reported in the paper, due to cost, we performed only a single run for each experiment. We batch sentences based on their input and approximate output token count for better GPU utilization. We set the total number of tokens to 700 for Mbart and 2000 for Bart models. We use gradient accumulation of 3 for Mbart and 9 for bart models boosting the effective batch size for better training.
Our models were trained on NVIDIA V100 GPU using the AWS platform. For a fair comparison, all models were trained for the same number of iterations of 50K.

  








\begin{table}\fontsize{8}{10}\selectfont
\centering
\begin{tabularx}{\linewidth}{l|c|c|c}
\toprule
\multirow{3}{*}{} & \multicolumn{3}{c}{\bf Dataset}  \\
\hline
\bf{} & {\bf RiSAWOZ} & {\bf CrossWOZ} & {\bf  MultiWOZ}\\
\hline
Languages & ZH, EN$^{*}$, DE$^{*}$ & ZH, EN & EN, ZH \\ 
\# Domains & 12  & 5 & 7  \\ 
\# Dialogues & 10,000  & 5,012 & 8,438 \\
\# Turns  & 134,580   & 84,692  & 115,424  \\
\# Slots  & 159  & 72 & 25 \\
\# Values & 4,061  & 7,871 & 4,510 \\
\bottomrule
\end{tabularx}
\caption{Statistics of the training datasets (excluding validation and test). $^{*}$Created using our methodology.}
\label{tab:train_stats}
\vspace{-1em}
\end{table}


\begin{table*}
\fontsize{9}{11}\selectfont
\centering
\scalebox{1}{
\begin{tabularx}{\linewidth}{l||c|c|c|c|c|r|r||r|r}
\toprule
 & \multicolumn{7}{c||}{SOTA} & \multicolumn{2}{c}{BART-CSP} \\
\hline
\multirow{2}{*}{Dataset} & \multirow{2}{*}{Model} & Context & Dialogue & Encodes & Predefined & \multirow{2}{*}{GJGA} & \multirow{2}{*}{JGA} & \multirow{2}{*}{GJGA} & \multirow{2}{*}{JGA} \\
 & & Encoder & History & State & Slots / Ont. & & & \\
\hline
RiSAWOZ  & MLCSG & Bi-GRU & Full & \ding{55} & \ding{51} & -- & 66.2  &  90.4 & \bf 76.9  \\
RiSAWOZ-EN-auto  & MLCSG & Bi-GRU & Full & \ding{55} & \ding{51} & -- & 36.2  &  88.8 & \bf 68.6  \\
\quad(-alignment)  & MLCSG  & Bi-GRU & Full & \ding{55} & \ding{51} & -- & 15.6 &   65.2 & \bf 22.9  \\
RiSAWOZ-DE-auto & MLCSG & Bi-GRU & Full & \ding{55} &  \ding{51} & -- & 13.4  &   86.7 & \bf 65.9 \\
\hline
CrossWOZ  & TRADE & Bi-GRU & Full & \ding{55} &  \ding{51} & 71.3$^{\dagger}$ & 36.1  & \bf 80.2 &  \bf 53.6  \\
CrossWOZ-EN & SOM & BERT & Partial & \ding{51} &  \ding{55} & -- & 32.3$^{\ddag}$  &  81.1 & \bf 52.3   \\
\hline
MultiWOZ 2.1 & MinTL & BART-large & Partial & \ding{51} &  \ding{55} & --  & \bf 53.6$^{\ddag}$  &  81.2 & \bf 53.7  \\
MultiWOZ 2.1 & STAR & BERT & Full & \ding{51} & \ding{51} & 78.7$^{*}$ & \bf 56.7  & \bf 81.2 & 53.7  \\
MultiWOZ-ZH 2.1  & SUMBT & BERT & Full & \ding{51} & \ding{51} & -- & \bf 46.0$^{\ddag}$  & 75.9 & \bf 46.3   \\
MultiWOZ 2.4 & STAR & BERT & Full & \ding{51} & \ding{51} & 90.2$^{*}$ & \bf  74.8  & \bf 91.7 & 70.4  \\
\bottomrule
\end{tabularx}
}
\caption{Comparison of results on test set for the BART-CSP model vs state-of-the-art on various datasets. The best result on JGA metric is in bold. GJGA is only bolded if SOTA is available. $^{\dagger}$RuleDST uses the previous system state and user dialogue acts as input. $^{*}$STAR-GT uses ground truth previous dialogue state as input. $^{\ddag}$ indicates results are reported from their paper. We reproduced all results for models that source code and pretrained models were available to ensure correct comparison. ``-align.'' denotes an ablation where translation is done without slot-value alignment.}
\label{tab:accuracies}
\vspace{-1em}
\end{table*}

\subsection{Metrics}
We evaluate the models using the following two metrics:
\begin{itemize}[leftmargin=*,noitemsep]
\item \textbf{Joint Goal Accuracy (JGA):} The standard metric of evaluation in DST is \textit{joint goal accuracy}, which measures the average accuracy of predicting all slot assignments to exact match (EM) for any given turn. 
To compute this metric for CSP, the belief state predicted in previous turn is used as input for the current turn. 


\item \textbf{Gold Joint Goal Accuracy (GJGA):} This metric is similar to JGA but is calculated on a turn by turn basis, with ground-truth belief state used as input. Assuming the belief state correctly captures the state up to current turn of the dialogue, this metric acts as an oracle in evaluation removing the compounding effect of errors from previous turns.
\end{itemize}


\section{Analysis and Results}

Table \ref{tab:accuracies} shows the results on test set for BART-CSP and previous SOTA models.
For a better comparison we have included some details for each model: 

\begin{itemize}[leftmargin=*,noitemsep]
    \item \textit{Context Encoder}: The neural model used to encode the input. 
    \item \textit{Dialogue History}: If dialogue history is included in the input. A turn is defined as a pair of user utterance and agent response. ``Partial'' history means only a few turns of dialogue are kept while ``Full'' history models encode all previous turns.
    \item \textit{Encodes State}: indicates if the user belief state up to the current turn is included in the input.
    \item \textit{Predefined Slots or Ontology}: whether the model design or the data processing step needs knowledge of slot names or values. 
\end{itemize}

As shown in Table~\ref{tab:accuracies}, all previous models encode either a partial or full history. BART-CSP encodes significantly less information as it relies only on the current turn and the latest belief state. This simplifies model design and improves data efficiency for training~\cite{yang2021comprehensive,kapelonis2022multi}.


Furthermore, models that rely on predefined ontologies require changes in architecture for new datasets. On the other hand, BART-CSP is a generative model that learns to copy slots from context and can be deployed for a new dataset as is.

 
On MultiWOZ, we report results for two models: MinTL which uses BART-large, and STAR\footnote{SOLOIST~\cite{peng2021soloist} and SCORE~\cite{yu2021score} achieve better performance than STAR; however, these models use additional dialogue datasets to either pre-train or fine-tune their models. To keep the comparison fair, we only consider prior work that trained on MultiWOZ only.} which uses BART as the context encoder. Our model outperforms MinTL despite using a smaller BART model, and achieves similar performance to STAR despite not having access to the full history. This shows bigger models do not necessarily yield better performance and model architecture and data representation are important too. 


\subsection{RiSAWOZ}
{\em The RiSAWOZ experiments shows that contextual semantic parsing delivers better accuracy than the state of the art on the original data sets; it is even more significant for translated data sets because it is more robust to translation errors.} 

BART-CSP improves the state of the art by 10.7\% on JGA to 76.9\% for the original Chinese data set. It provides an even greater improvement on the translated data sets: by 32.4\% to 68.6\% and by 42.5\% to 65.9\% on the automatically translated English and German data sets, respectively. 

BART-CSP holds two major advantages over models that predict the slot-value pairs from the dialogue history. First, by distilling the belief state into a concise representation, it reduces noise in the input that would otherwise be present in a long dialogue history. Second, by taking the belief state as input, the model becomes more robust to translation errors in utterances from previous turns than models that accept dialogue histories. Hence it is even more effective on translated datasets.

{\em Alignment is critical to generating a high-quality translated dialogue data set.} The English and German semantic parser show only a degradation of 8-11\% from the Chinese parser. In an ablation study of direct translation without alignment, the JGA on RiSAWOZ English drops from 68.6\% to 22.9\%, a difference of 45.7\%.

Alignment ensures that entities are translated to the right phrase from the target ontology. For example, the phrase ``\chin{姑苏区}" is translated to 
``Aguzhou district'' in a user utterance when the whole sentence is translated directly, but becomes ``Gusu district'' in the annotation. With alignment, both are translated identically to ``Gusu district.'' The correspondence of utterance and belief state leads to higher DST performance. 


\subsection{CrossWOZ}
Compared to MultiWOZ and RiSAWOZ, CrossWOZ is a more challenging dataset. Besides having longer conversations with more domains per dialogue, the cross-domain dependency is stronger. For example, the choice of location in one domain will affect the choice of location in a related domain later in the conversation, requiring models to have better contextual understanding.
CrossWOZ is not only a smaller, more complex dataset than RiSAWOZ, but also exhibits a higher misannotation rate, to be discussed in Section \ref{sec:errs}. 
The current state of the art result was obtained with TRADE, which achieves 36.1\% JGA. 

{\em The experiments with CrossWOZ also confirm that BART-CSP outperforms prior state-of-the-art models that encode full or partial history.} Specifically, it exhibits an improvement of 17.5\% in JGA on the original dataset and 20.0\% in JGA on the English translated data set. 

The GJGA metric for CrossWOZ was obtained by using RuleDST~\cite{zhu2020crosswoz}, a set of hand-written rules specialized to the dataset to compute the new belief state from the ground truth user and system dialogue acts. BART-CSP outperforms the use of RuleDST in GJGA by 9\%, showing that it is not necessary to handcraft these rules. 

The translated CrossWOZ-EN data have been manually corrected for slot-value errors. Application of our automatic slot-value alignment technique would have greatly reduced the tedious manual effort required.
In both GJGA and JGA, BART-CSP performs within 1\% of the original Chinese dataset on English CrossWOZ.

\subsection{MultiWOZ}


MultiWOZ is a challenging data set because of the well-documented problem of misannotations in the data set~\cite{eric2019multiwoz}. Misannotations teach the model to mispredict; conversely, correct predictions may be deemed to be incorrect. Thus the current state-of-the-art STAR model can only achieve an accuracy of 56.7\%. 

While BART-CSP accepts only the belief state as input context, the STAR model accepts both the previous belief state and the dialogue history. The latter offers an opportunity for the model to recover missing state values from the history, giving a 3\% advantage in JGA over BART-CSP. However, we note that once an agent misinterprets a user input, it is not meaningful to measure the accuracy for subsequent turns since the conversation would have diverged from the test data. 

On the other hand, parsing history has its own cost: (1) It is less data efficient as you need more data to learn the same task. (2) It requires a more complex model that can find relevant slots among a large number of sentences. BART-CSP outperforms STAR by a 2.5\% improvement in GJGA, suggesting that having the dialogue history as input can be detrimental when the past turns of a dialogue have been predicted correctly. On the Chinese translation of MultiWOZ, BART-CSP does slightly better than state of the art, improving JGA by 0.3\% to 46.3\%.

We also compare BART-CSP performance to MinTL, which is not SOTA, but uses more recent BART and T5 models as encoder. The results show that better performance on DST cannot be achieved by solely relying on better encoders with improved pretraining as both models are outperformed by STAR which uses BERT. 

Between MultiWOZ 2.1 and 2.4, BART-CSP results improve by 16.7\% on JGA and 10.5\% on GJGA, while STAR improves by 18.1\% on JGA and 11.5\% on GJGA, showing dependence of both BART-CSP and STAR on the quality of annotation. Because MultiWOZ 2.4 only corrects the validations and test sets, CSP is still affected by mis-annotations in the training dataset. 
The lack of an equally clean training set may be the reason BART-CSP does not exhibit as much improvement across the versions.

\section{Data and Error Analysis}
\label{sec:errs}

A manual inspection revealed the following sources of errors on the CSP model, showing some of the inference limitations and its susceptibility to misannotations. 

\subsection{Misannotations}
A substantial portion of incorrect predictions is due to existing annotation errors in all the datasets.
In particular, in a manual review of 200 randomly chosen turns from each dataset, RiSAWOZ exhibits a 10.0\% misannotation rate while CrossWOZ and MultiWOZ  exhibit 17.9\% and 26\% misannotation rates, respectively. 

 Prevalent misannotation error types observed in the three datasets are noted below, with examples in the appendix. 

\begin{itemize}[leftmargin=*,noitemsep]
    \item \textbf{Inconsistency:} Annotation inconsistency is a common issue with the Wizard-of-Oz data collection method. Examples of inconsistent annotations include inferred slots and slots that are mentioned by the agent but ignored by the user. 
    
    \item \textbf{Inexact Match:} Typos, i.e. minor mismatches between the utterances and the annotation slot values. Chinese is a homonym-heavy language. It is not unexpected for single-character mismatch typos to occur frequently in the dataset. A second kind of typo is for a character to be entirely missing in an entity name.
    \item \textbf{Missing Slots:} Sometimes, values for some slots  
    are simply just not included in the annotations. 
\end{itemize}

CrossWOZ and MultiWOZ 2.1 are also susceptible to the following:
\begin{itemize}[leftmargin=*,noitemsep]
    \item \textbf{Extra Slots:} The presence of slot names which are not mentioned by either the user or the agent. 
\end{itemize}

The following additional annotations problems are salient in MultiWOZ 2.1:
\begin{itemize}[leftmargin=*,noitemsep]
    \item \textbf{Delayed Annotations:} Slot values that are already confirmed by the user show up at a later turn in the conversation. 
\end{itemize}


In RiSAWOZ, the final parting turn in a dialogue has no annotations, indicating a state reset:
\begin{itemize}[leftmargin=*,noitemsep]
\item \textbf{Empty Annotation (Hard State Reset):} Some turns are missing annotations altogether. 
\end{itemize}

\subsection{Logical Relation Inference}
In RiSAWOZ, the model is expected to infer the logical relationships between entities. For instance, the price range for a \textit{fast food} restaurant should be \textit{cheap}; looking for an attraction to go to with a girlfriend implies the interest of a place where \textit{best-for-crowd} = ``lover's date''; similarly the desired hotel rating is to be inferred from the utterance rather than explicitly mentioned. 

For example:
\begin{itemize}[leftmargin=*,noitemsep]
    \item \chin{我们一家是外地的 ， 来苏州游玩 ， 你可以帮我找一个在吴中区 ， 中等消费水平的景点吗 } (Our family is foreign, we have come to Suzhou to have fun, could you help me find, in the WuZhong area, a medium priced attraction?)
    
    The slot value pair \chin{最适合人群 = ``家庭亲子''} (\textit{best-for-crowd} = ``family'') must be inferred.
    
    \item \chin{好的 。 再看看高新区有什么好吃的火锅店 ？}(Okay. Can you also see if GaoXin District has some good hotpot?)
    
    The slot \chin{价位 = ``偏贵''} (\textit{price} = ``expensive'') must be inferred from the fact that the food is hotpot.
\end{itemize}

We found that most models struggle with this type of inference which requires higher level language understanding and reasoning. It is unclear whether this weakness is inherent to the model or whether it is an artifact of inconsistent annotations. Of the misannotations counted, 33\% were missing the slot ``Best for crowd.'' 

\subsection{Expanded Range Inference}

In CrossWOZ, when a user requests, for instance, a restaurant where the cost per person is within a price range, oftentimes the agent cannot find such a restaurant, and responds with a suggestion with a cost outside of that range. In subsequent slots, the price range is expanded to include that cost, typically by rounding to the nearest ten. For example:

\begin{itemize}[leftmargin=*,noitemsep]
    \item \chin{你好，我想去吃饭。请帮我找一家人均消费是100-150元，有井冈山豆皮这个菜的餐馆。} (Hello, I'd like to go eat. Please help me find a place where the cost per person is 100-150 yuan, and has Jingangshan bean curd.) In this turn, the slot value for cost per person is \textit{100-150}. However, following the agent response, the slot changes: \chin{只为您找到一家叫西江美食舫(健德桥店)，但是它家的人均消费是83元。} (I could only find a place called XiJiang Gourmet Boat (Jiandeqiao location), but the cost per person is 83 yuan.) Since 83 is outside the original range \textit{100-150}, the slot value is expanded to \textit{80-150} for the remaining turns of the dialogue. 
\end{itemize}

We find that CSP struggles with such turns and will typically mispredict the slot value, assigning the previous range rather than inventing a new one.

\section{Conclusion}
\label{conclusion}


Given a dialogue dataset in one language, this paper shows how to build contextual semantic parsers for a new language using automatically machine-translated data. We propose an improved alignment approach for dialogues to ensure faithful translation of slot values. This removes the need for costly human-post editing used in all previous benchmarks.


We show that the compounding effects of translation noise across turns can be mitigated with a CSP model for dialogue state tracking. By leveraging pretrained seq2seq models such as BART, training with CSP can outperform state-of-the-art results on RiSAWOZ, CrossWOZ, and MultiWOZ-ZH, and remains competitive on MultiWOZ, despite not encoding any previous conversation turns or having access to a predefined ontology.



We use our methodology to create RiSAWOZ English and German, the first automatically created high-quality translated datasets for dialogue state tracking with no human in the loop. We have implemented our methodology as a toolkit\footnote{Code can be accessed at \url{https://github.com/stanford-oval/dialogues}} which developers can use to create a new multilingual dialogue dataset as well as a contextual semantic parser for it.

\section{Limitations}

Organic multilingual dialogue datasets (i.e. created without the use of translation) are scarce, which has limited the scope of our experiments. We would have liked to evaluate the generalization of our approach to other languages. For instance, we partially rely on machine translation models to create datasets. Available translation models for other language pairs, especially from/to low-resource languages have much lower quality, and it would be desirable to measure the effect of that in our experiments. 

Our methodology has only been applied to Human-to-Human dialogues annotated with slot-values. Although our approach is independent of data collection technique and formal representation, it should be applied and tested on datasets annotated with representations other than slot-values to study how well it can generalize. 

Previous studies~\cite{globalwoz,hung2022multi2woz} utilized human post-editing to guarantee the fluency and accuracy of the translated datasets. However, in order to reduce cost, we have decided not to use manual post-editing in this work. As a result, our findings could be an overestimation of the model's actual performance in real-world scenarios. In future research, we plan to rectify this by manually post-editing the validation and test portions of the datasets.

\section{Ethical Considerations}
\label{sec:ethics}

Our translation method replaces the manual work needed to create multilingual dialogue datasets usually done via crowdsourcing. Instead, it requires some computation time which can be an environmental concern. However, in practice, such additional computing is small and much cheaper than the cost of human annotation for the same amount of data.
The translation of the data set takes about half an hour on an Nvidia TITAN V GPU. Training takes about 6 hours on an Nvidia V100 GPU.
We did not use crowdworkers for this paper. The error analysis was done by the authors.

\section*{Acknowledgements}
This work is supported in part by the National Science Foundation under Grant No.~1900638, the Alfred P. Sloan Foundation under Grant No.~G-2020-13938, Microsoft, Stanford HAI, and the Verdant Foundation.

\bibliography{anthology,custom}
\bibliographystyle{acl_natbib}

\newpage
\appendix
\section{Appendix}

\subsection{Missannotation Examples}
See Table~\ref{tab:misann} below for missannotation examples discussed in Section~\ref{sec:errs}.

\begin{table*}
\newcolumntype{L}{>{\arraybackslash}m{4cm}}
\newcolumntype{M}{>{\arraybackslash}m{7cm}}
 \scalebox{0.7}{
\begin{tabular}{l|l|M|L|L}
\toprule
\bf{Error Type}  & \bf{Dataset} & {\bf Agent Utterance; User Utterance} & {\bf Annotation} & {\bf Correct Annotation}\\
\hline
\multirow{2}{*}{Delayed Annotation} & \multirow{2}{*}{MultiWOZ} & great , i can get you a ticket for that train . how many people are riding with you ?; i need to book it for 6 people , can i get the reference number too ?& train day = " thursday " train departure = " cambridge " train destination = " birmingham new street " train leaveat = " 10:00 " & \textbf{train book people = " 6 "} train day = " thursday " train departure = " cambridge " train destination = " birmingham new street " train leaveat = " 10:00 "\\
&& [next turn]  can i confirm you want to book this train for 6 people ?; yes , i would like to book the train for 6 people . i need the reference number , please . & \textbf{train book people = " 6 "} train day = " thursday " train departure = " cambridge " train destination = " birmingham new street " train leaveat = " 10:00 " & (already correct)\\
\hline
Extra Slot & MultiWOZ & ; hello ! i am planning my trip there and i am trying to find out about an attraction called kettle s yard . what can you tell me about it ?
& attraction name = " kettles yard " , \textbf{attraction area = " west "} & attraction name = " kettles yard "\\
\hline
\multirow{2}{*}{Empty Annotation} & \multirow{2}{*}{RiSAWOZ} & \chin{;你好 ， 刚到苏州 ， 想先找个餐厅吃点东西 。 有价位便宜的江浙菜餐厅吗 }& \textbf{ null }&  \chin{餐厅 价位 = " 便宜 " 餐厅 菜系 = " 江浙菜 "} \\
&& Hello, I just came to Suzhou and am looking for a restaurant to grab a bite. Is there a cheap Jiangzhe restaurant?& & Restaurant Price = "Cheap" Restaurant Cuisine = "Jiangzhe" \\
\hline
\multirow{2}{*}{Inexact Match} & \multirow{2}{*}{RiSAWOZ} & \chin{;你好 ， 我这几天在苏州度假 ， 明天准备去狐狸家手工奶酪这家餐厅吃饭 ， 但不是很了解 ， 你能帮我查查那个餐馆附近有没有地铁能直达呢 ？}& \chin{餐厅 名称 = " 狐狸家手工\textbf{酸}奶酪 "} & \chin{餐厅 名称 = " 狐狸家手工奶酪 "}\\
&& ; Hello, I'm vacationing in Suzhou these several days, tomorrow I plan to go to the Fox Family Handmade Cheese restaurant to eat, but I don't really understand, can you help me look up whether there is direct subway access to anywhere near there? & Restaurant Name = "Fox Family Handmade \textbf{Yogurt} (Yogurt = Sour Cheese)"& Restaurant Name = "Fox Family Handmade Cheese" \\
\hline
\multirow{2}{*}{Missing Slot} & \multirow{2}{*}{RiSAWOZ} & \chin{有呀 ， 推荐您去鑫花溪牛肉米粉 。}; \chin{这家店地址在哪 ？} & \chin{餐厅 价位 = " 中等 " 餐厅 菜系 = " 快餐简餐 "} &  \chin{餐厅 价位 = " 中等 " \textbf{餐厅 名称 = " 鑫花溪牛肉米粉 "} 餐厅 菜系 = " 快餐简餐 "}\\
&& Yes, I recommend you go to Xinhuaxi Beef Noodle; What is the address of this place? & Restaurant Price = "Medium" Restuarant Cuisine = "Quick and Easy"& Restaurant Price = "Medium" \textbf{Restaurant Name = "Xinhuaxi Beef Noodle"} Restaurant Cuisine = "Quick and Easy" \\
\bottomrule
\end{tabular}
}
\caption{Prevalent annotation error types found in the datasets.}
\label{tab:misann}
\end{table*}

\end{document}